\documentclass[journal, letterpaper]{IEEEtran}

\IEEEoverridecommandlockouts
\usepackage[letterpaper, left=0.75in, right=0.75in, bottom=0.75in, top=0.67in]{geometry}

\usepackage{cite}

\usepackage{url}

\usepackage{amsmath,amssymb,amsfonts}
\usepackage{algorithmic}
\usepackage{graphicx}
\graphicspath{ {./images/} }

\usepackage{tabularx, booktabs}
\usepackage{multirow}
\usepackage[footnotesize]{caption}

\usepackage{dirtytalk}

\usepackage{algorithm}
\usepackage{footnote}

\usepackage[caption=false, font=footnotesize]{subfig}

\usepackage{etoolbox}

\usepackage{gensymb}
\usepackage{textcomp}
\usepackage{xcolor}

\def\BibTeX{{\rm B\kern-.05em{\sc i\kern-.025em b}\kern-.08em
    T\kern-.1667em\lower.7ex\hbox{E}\kern-.125emX}}
\begin{document}

\pagenumbering{gobble}
\newgeometry{left=0.75in, right=0.75in, bottom=0.75in, top=0.95in}

\title{\LARGE \bf CNN-Based Camera Pose Estimation and Localisation of Scan Images for Aircraft Visual Inspection\\
}

\author{Xueyan Oh$^{1}$, Leonard Loh$^{1}$, Shaohui Foong$^{1}$, Zhong Bao Andy Koh$^{2}$, Kow Leong Ng$^{2}$, Poh Kang Tan$^{2}$, Pei Lin Pearlin Toh$^{2}$, and U-Xuan Tan$^{1}$

\thanks{$^{1}$X. Oh, L. Loh, S. Foong and U-X. Tan are with the Pillar of Engineering Product Development, Singapore University of Technology and Design, Singapore,
        {\tt\small xueyan\_oh@mymail.sutd.edu.sg}, \tt\small uxuan\_tan@sutd.edu.sg.}
\thanks{$^{2}$Z. B. A. Koh, K. L. Ng, P. K. Tan and P. L. P. Toh are with ST Engineering Aerospace Ltd., Singapore}
\thanks{A preliminary version \cite{CPE_Aircraft} of this paper has been published in ICRA 2021.}

}

\makeatletter
\patchcmd{\@maketitle}
  {\addvspace{0.5\baselineskip}\egroup}
  {\addvspace{-2\baselineskip}\egroup}
  {}
  {}
\makeatother

\maketitle

\begin{abstract}
General Visual Inspection is a manual inspection process regularly used to detect and localise obvious damage on the exterior of commercial aircraft. There has been increasing demand to perform this process at the boarding gate to minimise the downtime of the aircraft and automating this process is desired to reduce the reliance on human labour. Automating this typically requires estimating a camera’s pose with respect to the aircraft for initialisation but most existing localisation methods require infrastructure, which is very challenging in uncontrolled outdoor environments and within the limited turnover time (approximately 2 hours) on an airport tarmac. Additionally, many airlines and airports do not allow contact with the aircraft's surface or using UAVs for inspection between flights, and restrict access to commercial aircraft. Hence, this paper proposes an on-site method that is infrastructure-free and easy to deploy for estimating a pan-tilt-zoom camera's pose and localising scan images. This method initialises using the same pan-tilt-zoom camera used for the inspection task by utilising a Deep Convolutional Neural Network fine-tuned on only synthetic images to predict its own pose. We apply domain randomisation to generate the dataset for fine-tuning the network and modify its loss function by leveraging aircraft geometry to improve accuracy. We also propose a workflow for initialisation, scan path planning, and precise localisation of images captured from a pan-tilt-zoom camera. We evaluate and demonstrate our approach through experiments with real aircraft, achieving root-mean-square camera pose estimation errors of less than 0.24 m and 2{\degree} for all real scenes.
\end{abstract}


\vspace{-3mm}

\section{Introduction}
General Visual Inspection (GVI) is a widely used technique as part of regular inspections of aircraft such as during pre-flight inspections on an airport tarmac or during maintenance usually performed in a hanger. This process involves visual examinations of the aircraft’s exterior for noticeable damage or irregularities and provides a means for early detection of typical air-frame defects \cite{R1}. Currently, this is manually performed by well-trained personnel, which is labour intensive and have high error rates \cite{R1,R2}. To address this, research has shown that automating this process can increase the speed of inspection, and increase the accuracy of detecting defects among other benefits \cite{R5,R41}.

Many studies have explored automating the task of detecting defects on the surface of aircraft. Mumtaz \textit{et al.} \cite{R46} propose a visual approach using directional energies of textures to detect and distinguish between cracks and scratches. Malekzadeh \textit{et al.} \cite{R4} train deep neural networks to detect defects within images of an aircraft fuselage. Miranda \textit{et al.} \cite{R44} focus on detecting and inspecting exterior screws on aircraft by training a deep convolutional neural network (DCNN) to extract screws from images and apply algorithms to evaluate the state of each screw. Miranda \textit{et al.} \cite{R45} address the issue of data imbalance between different defect categories in datasets used to train defect classification networks by proposing a hybrid machine learning method that improves defect classification performance for the under-represented category. Doğru \textit{et al.} \cite{R40} tackle the issue of small datasets for aircraft defects by using a DCNN combined with data augmentation techniques to improve the performance of detecting dents on aircraft surfaces within images. While the above studies show the potential of automating defect detection within images, localisation of the defects with respect to the aircraft is still missing.

\begin{figure}[t]
\centerline{\includegraphics[width=\columnwidth]{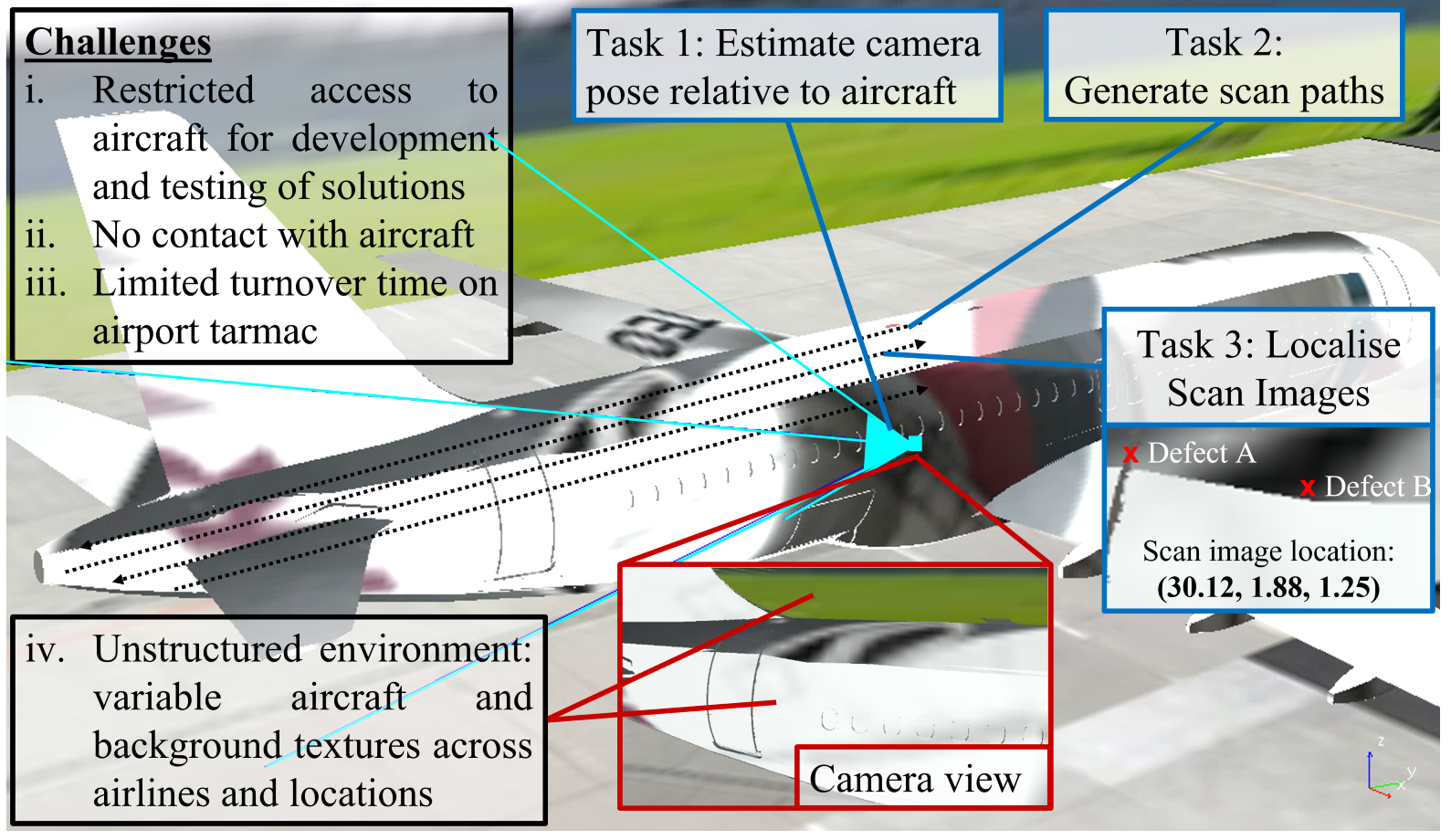}}
\vspace{-2mm}
\caption{We propose an infrastructure-free and contact-less method to estimate a PTZ camera's pose with respect to an aircraft and generate a scan path to obtain scan images labelled with their location on the aircraft's surface.}
\label{fig:1}
\vspace{-6mm}
\end{figure}

On top of detecting abnormalities, other studies have suggested automation of the tedious data collection process with the use of ground-based robotic systems. Several works \cite{R5, R42} study the use of a Pan-Tilt-Zoom (PTZ) camera mounted on an autonomous mobile ground robot that relies on navigating to pre-defined checkpoints with the help of laser range finders or image data. However, these systems are only designed to inspect larger-scale features from the ground, such as an air-inlet vent or an oxygen bay handle, and do not require high camera viewpoint accuracy. Jovančević \textit{et al.} \cite{R41} propose the use of a similar autonomous mobile ground robot that moves along a pre-defined path around an A320 within a hanger to collect 3D point cloud data using a 3D scanner. This work focuses on the analysis of point cloud data for defect detection and classification and not on localising them. These solutions are also designed to be implemented in a hangar during maintenance, which provides a controlled environment to set up infrastructure for establishing checkpoints around the aircraft. Hence, they are not suitable to be performed outdoors on the airport tarmac and with limited turnover time (approximately 2 hours) between flights.

The use of Unmanned Aerial Vehicles (UAV) have also been explored for aircraft inspection. Malandrakis \textit{et al.} \cite{R43} and Tzitzilonis \textit{et al.} \cite{R48} use a commercially available UAV attached with a UV LED torch and a camera for autonomous inspection of an aircraft wing. Both works suggest that the UAV autonomously navigate between fixed waypoints along the wing using optical flow and capture images. However, the initial pose of the UAV is still required. Umberto and Salvatore \cite{R1} also use a UAV to automate defect detection and maintains a fixed vertical distance from the aircraft surface using ultrasonic sensors but is piloted by a human operator and relative pose to the aircraft is not known. Cazzato \textit{et al.} \cite{R39} estimate the pose of a UAV relative to an aircraft by comparing the features of known landmarks between the real image and a reference 3D model. However, the landmarks are specific logos and flags painted on the aircraft and assumed to be present on the reference model. This is not feasible in reality as the paintwork of an aircraft changes from time to time and differs between airlines. Lastly, Bugaj \textit{et al.} \cite{R38} present a methodology and the benefits for using semi-autonomous UAVs in aircraft pre-flight inspections in a safe manner. While these studies show the potential of using UAVs in aircraft inspections, none of them explore an end-to-end solution that can be quickly deployed in uncontrolled environment and capable of precisely localising detected defects, and the use of UAVs is also prohibited in many airports due to safety and security concerns.

It is difficult to localise a detected defect just by looking at an optically zoomed in image as there are often no location-specific features within an image. Additionally, many airlines and airports do not allow contact with the aircraft's surface between flights, causing pose estimation methods that require physically attaching sensors or markers onto the surface of aircraft to be unusable. To obtain an image's location on the aircraft's surface, it is required to first determine the camera’s initial pose relative to the aircraft which can be referred to as Camera Pose Estimation (CPE). The recent increase in the use of Deep Convolutional Neural Networks (DCNNs) for monocular CPE has shown its potential to out-perform classical 3D structure-based methods in several aspects, including shorter inference times, smaller memory \cite{R9}, and robustness to uncontrolled environments \cite{R10}. This shows the potential viability of applying DCNNs for CPE as an infrastructure-free and contact-less method to initialise aircraft inspection systems on airport tarmacs, under limitations such as prohibited contact with aircraft, time constraints and variations in background and lighting.

Our main contributions in this paper are as follows:

\begin{itemize}

\item We propose an end-to-end workflow for obtaining scan images labelled with their location on an aircraft's surface using an easily deployable PTZ camera system.

\item We propose a method to initialise and estimate the pose of the camera used for inspection without requiring physical contact with or prior access to a real aircraft by using a DCNN fine-tuned on only synthetic images of a 3D aircraft model to estimate the camera's pose given a real image as input.

\item We improve the network’s performance by adding a new component to the loss function. This additional loss component leverages on known geometry of an aircraft to provide a geometric relationship between the predicted position and orientation of the camera.

\end{itemize}

In this work, we build on the preliminary version \cite{CPE_Aircraft} of this paper and the key differences include: 1) proposed approach of end-to-end workflow from initialisation of the inspection camera to localising scan images; 2) experimental validation of the proposed algorithm with aircraft in different scenes with more comprehensive evaluation of our pose estimation method; 3) a scanning strategy for a PTZ camera to scan and label images of the upper-half surface of an aircraft with their 3D coordinates; and 4) demonstration of the complete workflow on a real aircraft.

\section{Related Work}

\subsection{CNN-Based Camera Pose Estimation} \label{cnncpe}
CPE can be described as using an input image to output an estimate of the pose – position and orientation – of the camera \cite{R9}. In most cases, the pose is obtained with respect to a predefined global reference. PoseNet \cite{R11} and other similar deep architectures that predict a camera pose \cite{R13,R15,R18} share a common process, where images from a database are used as input into a DCNN for training, with the aim of minimising the error between the predicted and ground truth pose. PoseNet \cite{R11} is the pioneer to introduce the use of a DCNN – based on a modified GoogLeNet \cite{R19} – to directly regress a camera pose, and many improved methods based on deep learning architectures have since been proposed \cite{R9}. These deep pose estimation methods have been fine-tuned and tested on publicly available indoor and outdoor datasets that have been collected using hand-held camera devices and using software to automate the retrieval of camera poses \cite{R11}. However, models fine-tuned on these datasets may not accurately represent their effectiveness in CPE for our application due to the difficulties in obtaining a real dataset to train on in the first place, as well as the need to accommodate scene changes such as different appearance of the same aircraft model and its background.

\subsection{Using Synthetic Images to Fine-Tune DCNNs} \label{syncnncpe}
Several research \cite{R24,R26,rnncpe,edgeseg} explore fine-tuning DCNNs without the need for real images by using synthetic scenes for tasks related to camera pose estimation. Among these, Sadeghi and Levine \cite{R24} propose a learning method for a drone to autonomously navigate an indoor environment without collision, by training a network through reinforcement learning using only 3D CAD models. Only RGB images rendered from a 3D indoor environment with random textures, object positions and lighting are used to train the CNN to output velocity commands, and they achieve autonomous drone navigation and obstacle avoidance. While they explore the ability of a network trained on synthetic images to generalise to the real world, their objective is to avoid collision as opposed to CPE.

Acharya \textit{et al.} \cite{R26} propose a solution for indoor CPE by fine-tuning PoseNet \cite{R11} using synthetic images rendered from a low-detail 3D indoor environment, modelled with reference to a Building Information Model (BIM). This work explores different methods of rendering, from cartoon-like to photo-realistic and textured to rendering only edges within each scene, achieving CPE from real images with an accuracy of about 2 m. Another study \cite{rnncpe} proposes to use synthetic image sequences to fine-tune a recurrent neural network for CPE and show that it improves CPE accuracy when testing on real image sequences, achieving CPE accuracy of about 1.6 m. More recently, Acharya \textit{et al.} \cite{edgeseg} show that combining edge maps and semantic segmentation can close the domain gap between synthetic images for training the CNN and real images and improve sim-to-real camera pose estimation, achieving an accuracy of 1.12 m and 6.06{\degree} in indoor scenes. However, these methods have only been tested in environments with substantial changes in scenes and viewpoints as the camera relocates within the environment. This is as opposed to differentiating the camera pose between images that are captured with slight changes in viewpoint in the context of aircraft GVI. Moreover, the accuracy reported in most of these works are insufficient for our application.

\subsection{Generating Synthetic Images via Domain Randomisation} \label{domainrandomisation}
Tobin \textit{et al.} \cite{R25} investigate the use of domain randomisation to bridge the gap between simulation and reality. They argue that models fine-tuned with only synthetic scenes can generalise to real scenes if the scenes are diverse enough. The authors generate their dataset by uniformly randomising many aspects of their domain, including size, shape, position and colour of objects in each scene, and successfully teach a robotic arm to pick objects within a real, crowded indoor environment using only “low-fidelity” rendered images. Following this, others \cite{R27,R28,R29} also use domain randomisation for deep pose estimation tasks without training on real data. Despite being robust to object distractors, these models only apply to object pose estimation and often have other unchanging major objects such as a table where objects are placed on or a robot gripper which provides useful information of each object’s pose relative to these major objects in the scene. We also use domain randomisation in our work, but focus on CPE with respect to an aircraft without any other known objects in the scene.

In summary, it is challenging to develop deep learning based solutions for CPE with respect to an aircraft on an airport tarmac due to the limited access to real aircraft and solutions need to be robust to large variations in environment and aircraft texture across airlines. We propose to address this challenge by eliminating the need to obtain real images for fine-tuning and use only synthetic images of the aircraft’s 3D model in scenes varied using domain randomisation. We also leverage on known geometry of an aircraft’s surface from its Structural Repair Manual (SRM) to explore a geometric relationship between the camera's position and orientation within the network's loss function and improve the pose estimation accuracy. To complete the workflow, we propose a path planning method for automated scanning of an aircraft with a PTZ camera that uses the output of CPE as well as information from the 3D model of the aircraft and label every captured image with the estimated location of its centre pixel on the aircraft surface.

\begin{figure*}[t]
\centering
\centerline{\includegraphics[width=0.8\textwidth]{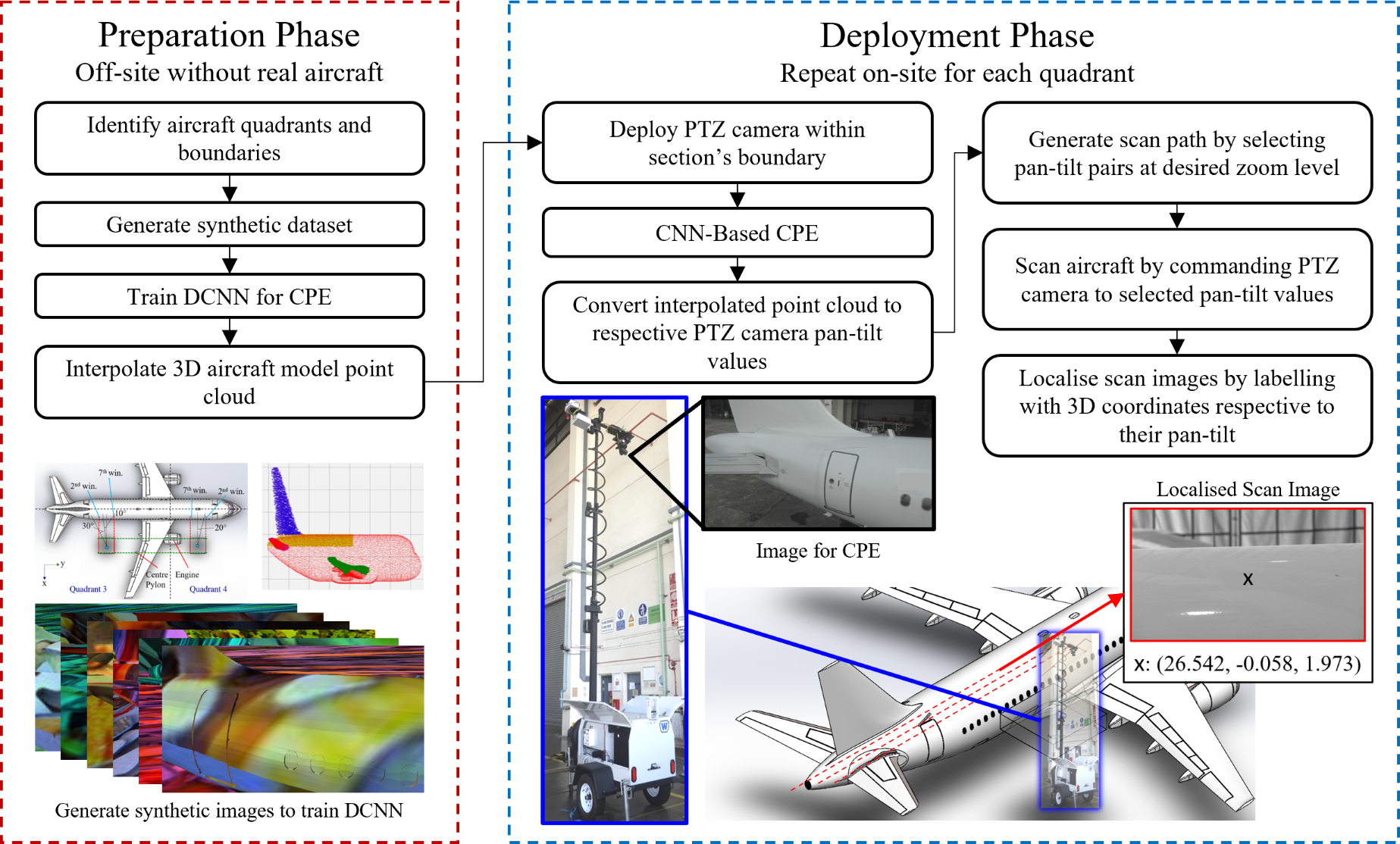}}
\vspace{-1mm}
\caption{Overview of our proposed workflow for scanning and localising images for aircraft visual inspection on an airport tarmac between flights. Our method is easy to deploy and does not require additional infrastructure or attaching sensors or markers onto the aircraft for camera pose estimation (CPE).}
\label{fig:workflow}
\vspace{-5mm}
\end{figure*}

\section{Workflow for Scanning and Localising Images for Aircraft Visual Inspection}\label{workflow}

Performing automated visual inspection of aircraft between flights demands minimal infrastructure due to the short turn-over time, and physical contact with the aircraft is prohibited by some airlines and airports. We propose a workflow, illustrated in Fig.~\ref{fig:workflow}, that achieves this using only the same PTZ camera for inspection mounted onto an extendable mast. The workflow consists of two phases, the first is called the \emph{Preparation Phase} and is only required when preparing the system for a new aircraft model or to improve the performance of the system (such as to fine-tune the DCNN). The second phase is called the \emph{Deployment Phase} which can be repeated on aircraft of the same model as long as the \emph{Preparation Phase} has been completed for that model. 

The \emph{Preparation Phase} starts by sectioning the chosen aircraft model (we choose an Airbus A320 in this work) into four quadrants and identifying their boundaries (both position and orientation) in which operators can reasonably position the PTZ camera (on its mast) within using aircraft features as visual guides. Second, a fully synthetic dataset is separately generated for each quadrant from a simulator and used to fine-tune a DCNN to predict the pose of a camera relative to the aircraft given a single image. Third, the 3D aircraft model's point cloud is interpolated to prepare for scan path generation during the \emph{Deployment Phase}.

In the \emph{Deployment Phase}, a PTZ camera of known FOV is set up within the proposed boundary of a chosen quadrant to be scanned. The PTZ camera is first initialised with CNN-based CPE using an image captured by the PTZ camera at full FOV. Next, points in the interpolated 3D aircraft model point cloud from the \emph{Preparation Phase} is converted to their respective PTZ camera pan-tilt values, given the estimated pose of the camera with respect to the aircraft. A scan path is then generated by selecting pan-tilt pairs based on the desired zoom level for the scan. Lastly, the scan is performed by commanding the PTZ camera to each of the selected pan-tilt pairs at the desired zoom level to capture images and label their centre pixels with their respective Cartesian coordinates.

\begin{figure}[t]
\centerline{\includegraphics[width=\columnwidth]{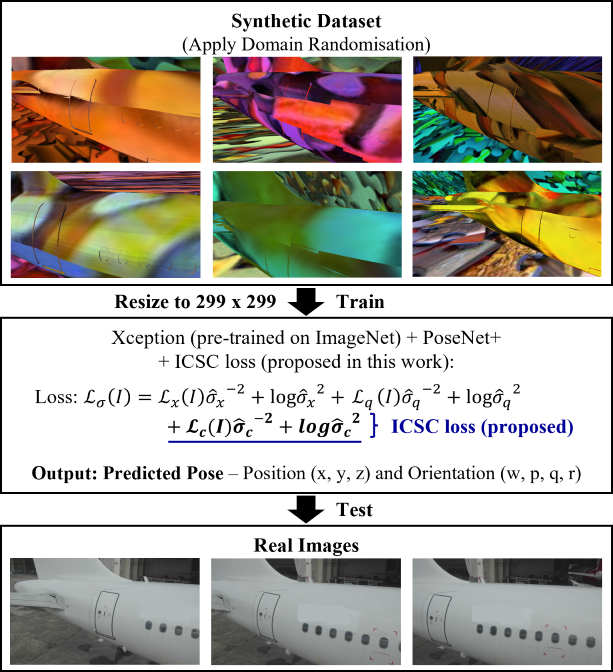}}
\vspace{-1mm}
\caption{Summary of deep learning approach.}
\label{fig:2}
\vspace{-7mm}
\end{figure}

\section{CNN-Based Camera Pose Estimation} \label{cnn-based_cpe}
We first design a setup with realistic constraints and assumptions for how a PTZ camera of known specifications can be easily deployed next to an aircraft for the purpose of inspecting the upper surface of the aircraft. More information of the setup is mentioned in \ref{proposed_setup}. Based on this setup, we use a virtual camera within a virtual 3D environment to capture images while applying domain randomisation to generate our synthetic dataset. We use this synthetic dataset to fine-tune a network that regresses the camera’s pose from an input image. We base our network on an existing PoseNet variant with learnable weights \cite{R18} (we refer to as PoseNet+). In addition, we propose modifying its loss function by introducing an additional component that provides a geometric relationship between the camera’s position and orientation. Fig.~\ref{fig:2} summarises our deep learning approach.

\subsection{Proposed Setup with PTZ Camera} \label{proposed_setup}

\begin{figure}[t]
\centerline{\includegraphics[width=\columnwidth]{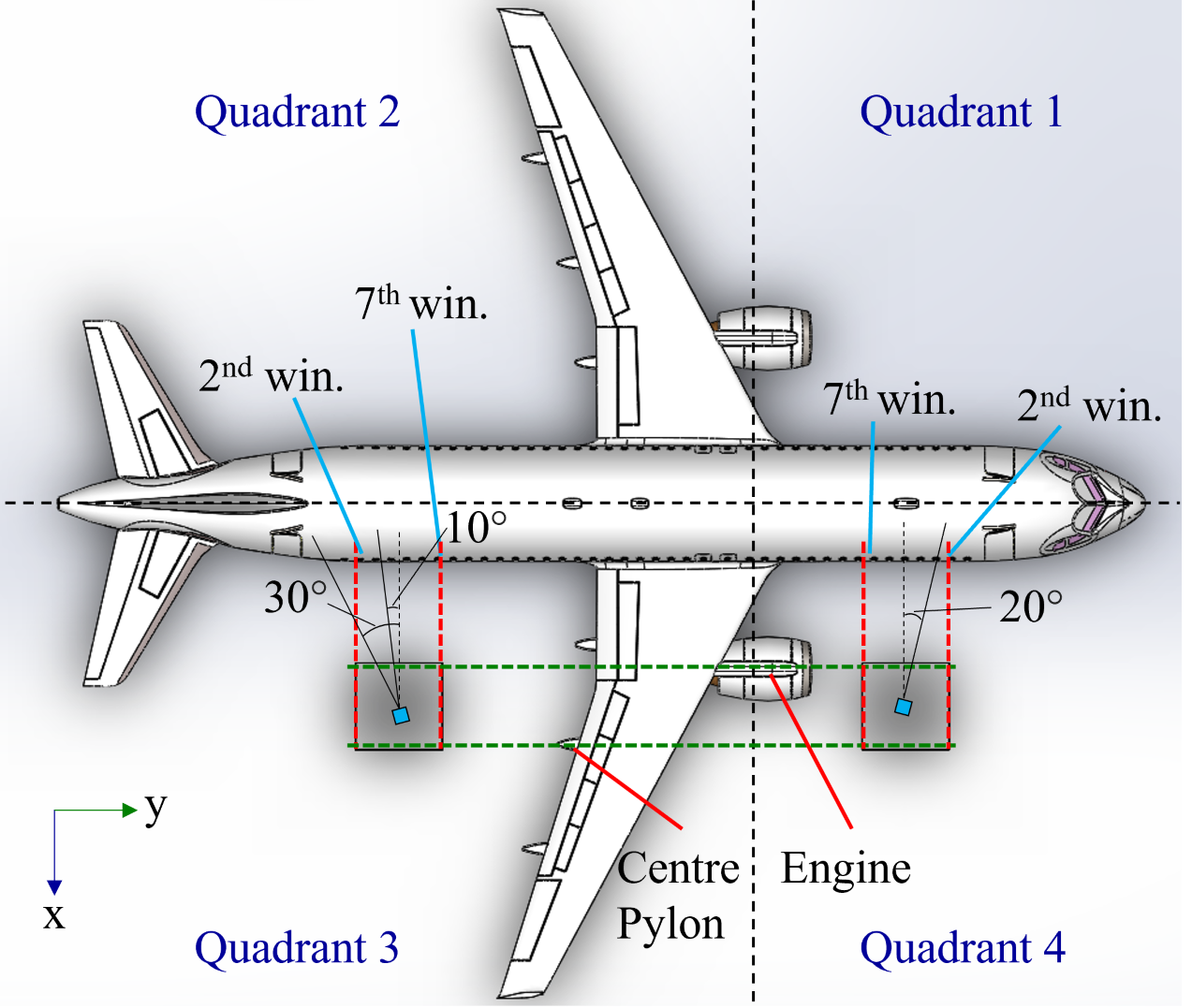}}
\vspace{-2mm}
\caption{Aircraft quadrants and proposed boundary for PTZ camera position and axes direction. Windows and pylons of an A320 are used as a visual guide to position the PTZ camera within the proposed boundary.}
\label{fig:3}
\vspace{-3mm}
\end{figure}

\begin{figure}[t]
\centerline{\includegraphics[width=\columnwidth]{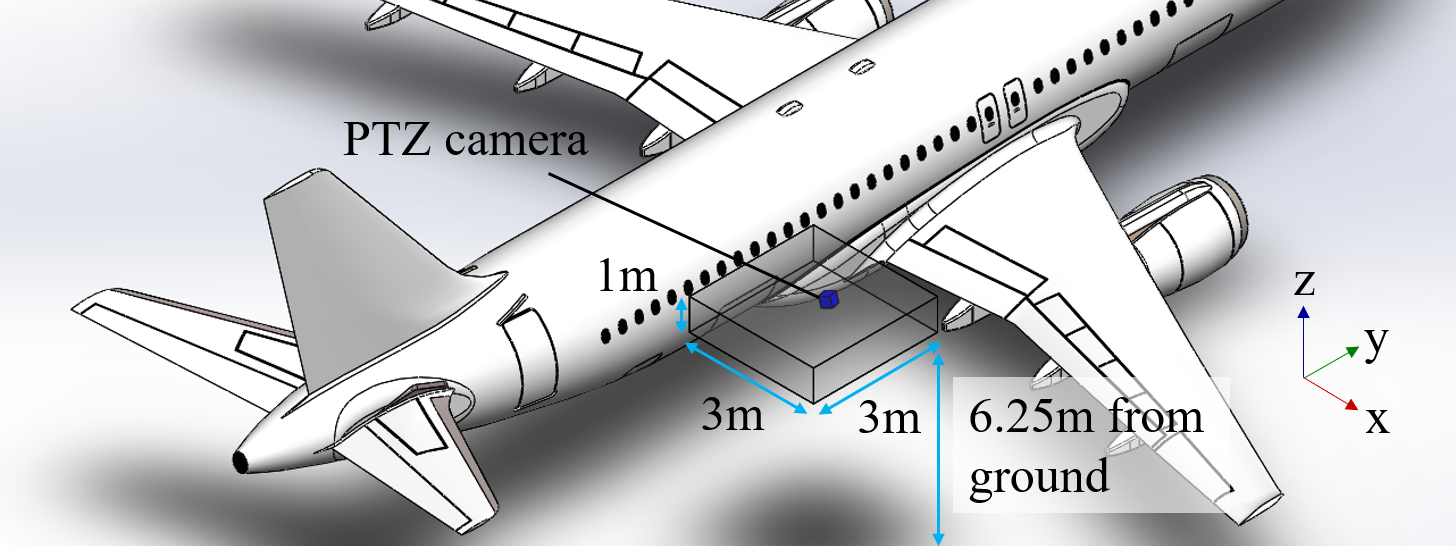}}
\vspace{-2mm}
\caption{Proposed 3D boundary for PTZ camera position in quadrant 3 with direction of axes.}
\label{fig:4}
\vspace{-4mm}
\end{figure}

We split an aircraft into four quadrants (refer to Fig.~\ref{fig:3}) and use only one side, quadrants 3 and 4, of an Airbus A320 (A320 in short) to illustrate our method and assume that the other half of the aircraft (quadrants 1 and 2) is similar and the method can be easily mirrored. It is assumed that the PTZ camera's specifications are known and we use its maximum FOV (at 1x zoom) for initialisation. It is also assumed that the PTZ camera’s base can be easily levelled (no roll and pitch relative to the ground) with the use of a gimbal or level gauge, reducing the problem to 4 Degrees of Freedom - position and yaw. The following steps are proposed for deployment:
\begin{enumerate}
\item Position the PTZ camera within a reasonable region as illustrated in Fig.~\ref{fig:3} and Fig.~\ref{fig:4}, which includes an area of 3 m by 3 m in each quadrant, and can be easily performed using visual check.
\item Raise the PTZ camera to a height of 6.25 m to 7.25 m from the ground via equipment such as an electronic mast or a boom lift and easily approximated with the use of accessible equipment such as a range finder. 
\item Rotate the camera's base about the z-axis to face the aircraft perpendicularly (within ${\pm} 10${\degree} yaw error).
\item From its home pan-tilt orientation, pan the camera 20{\degree} towards the aircraft’s tail (quadrant 3), or 10{\degree} towards the aircraft’s nose (quadrant 4), and tilt 18{\degree} towards the ground via software commands and capture an image for initialisation.
\end{enumerate}

Fig.~\ref{fig:3} shows the features (windows, pylon, and engine) of an A320 to use as visual guides for the quick manual positioning of the mast, while Fig.~\ref{fig:4} illustrates the proposed permissible 3D space where the PTZ camera can be set up using quadrant 3 as an example. The allowable yaw error (referred to as \begin{math}\alpha\end{math}) of ${\pm} 10${\degree} due to manual orientation towards the aircraft suggests that the camera is oriented to between +10{\degree} and +30{\degree} about the z-axis for quadrant 3, and between 0{\degree} and -20{\degree} about the z-axis for quadrant 4.


\subsection{Virtual Environment and Synthetic Dataset}
\begin{figure}[t]
\centerline{\includegraphics[width=\columnwidth]{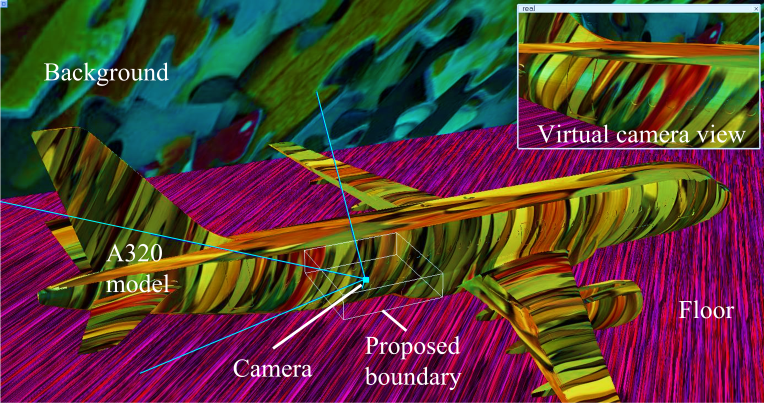}}
\vspace{-1mm}
\caption{Virtual setup with an instance of domain randomisation.}
\label{fig:5}
\vspace{-5mm}
\end{figure}

We obtain the 3D model of an A320 from a GrabCAD contribution \cite{R31}. Minor modifications are made to match general features and overall dimensions of a real A320, based on details obtained from an A320’s SRM. Our virtual setup is shown in Fig.~\ref{fig:5}. To create this 3D environment, we place our 3D model into a scene in robot simulator CoppeliaSim \cite{R33}. A large wall is added on one side of the aircraft model as background. A virtual camera is placed beside the aircraft and its FOV is set to match the real PTZ camera at 1x zoom.

We apply domain randomisation \cite{R25} when generating our synthetic dataset as it has been demonstrated to be capable of generalising to real-world data, given sufficient simulated variability. We use a free stock image of a randomly scattered puzzle to apply as texture for the ground, aircraft model, and background. We randomise the following aspects when capturing each image for our dataset and use quadrant 3 to illustrate our approach:

\begin{itemize}
\item The PTZ camera’s position within the proposed 3 m x 3 m x 1 m boundary in quadrant 3;
\item The PTZ camera’s pan between +10{\degree} and +30{\degree};
\item The PTZ camera’s tilt between -17.5{\degree} to -18.5{\degree}. (slight tolerance of ${\pm} 0.5${\degree} from the proposed 18{\degree} tilt);
\item Colour – RGB values of both ambient and specular components for the texture of every object; and
\item The position, orientation, as well as horizontal and vertical scaling factors of textures applied onto all surfaces.
\end{itemize}
For each camera with a different FOV at 1x zoom, we generate 4000 synthetic images for fine-tuning, 700 for validation and 300 for testing.

\subsection{Deep Learning Approach}
We use an improved variant \cite{R18} (which we refer to as PoseNet+) of PoseNet \cite{R11} for its ability to directly regress for CPE given a single input image. We apply PoseNet+'s approach onto a more recent deep architecture, Xception \cite{R35}, as it results in substantially better performance than GoogLeNet (Inception v1) originally used in PoseNet+. We modify the Xception network in a similar fashion to PoseNet+, by replacing the softmax layer with regression layers that output position (x, y, z) and orientation vectors (quaternions – w, p, q, r). We resize every input image to match the network's 299 x 299 pixel input size without a centre crop as we find this to improve performance and attribute this to the increase in features and other spatial information that may be present in the whole image despite the distortion from resizing.

PoseNet+'s loss function learns a weighting between the position and orientation components and is formulated using the concept of homoscedastic uncertainty - a measure of uncertainty of the task and is independent of input data \cite{R34}. It is defined as:
\begin{equation}
\mathcal{L}_\sigma(I) =  \mathcal{L}_x(I)\hat{\sigma}^{-2}_x + \log\hat{\sigma}^2_x + \mathcal{L}_q(I)\hat{\sigma}^{-2}_q + \log\hat{\sigma}^2_q
\label{eq:2}
\end{equation}
where \begin{math}\mathcal{L}_x(I) = \|x-\hat{x}\|_2\end{math} and \begin{math}\mathcal{L}_q(I) = \|q-\frac{\hat{q}}{\|\hat{q}\|}\|_2\end{math}, with \begin{math}\hat{x}\end{math} and \begin{math}\hat{q}\end{math} representing the predicted position and orientation vectors respectively while \begin{math}x\end{math} and \begin{math}q\end{math} represent ground truth pose. \begin{math}\hat{\sigma}^2_x\end{math} and \begin{math}\hat{\sigma}^2_q\end{math} represent the homoscedastic uncertainties and are optimised with respect to the loss function through back propagation, and the reason for their usage is described in the paper introducing PoseNet+ \cite{R18}.

Kendall \textit{et al.} \cite{R18} also explore removing the need for balancing positional and rotational loss weights by directly learning from geometric reprojection loss but find that this loss was unable to converge without first pretraining on their originally proposed loss function \eqref{eq:2}. Inspired by their use of scene geometry, we propose to modify loss function \eqref{eq:2} by introducing an additional loss component \begin{math}c\end{math} that uses the scene coordinate of each image’s centre pixel instead of reprojection. This is obtained by finding the point of intersection between the equation describing the camera’s optical axis (as a function of \begin{math}x\end{math} and \begin{math}q\end{math}) and the aircraft's surface. With our proposed setup, we find this point of intersection is always on the upper half of the fuselage and propose to model the aircraft's surface as the equation of a cylinder. With \begin{math}c\end{math} as the Cartesian coordinates of any point on the cylinder's surface, the equation of the surface is given by:
\begin{equation}
c_x^2 + (c_z-h_0)^2 = r_0^2
\label{eq:4}
\end{equation}Where \begin{math}c_x\end{math}, \begin{math}c_y\end{math} (any value along the cylinder’s length) and \begin{math}c_z\end{math} are the coordinates of a point on the cylinder's surface, \begin{math}h_0\end{math} is the displacement of the cylinder’s cross-sectional centre from the scene’s origin, and \begin{math}r_0\end{math} is the aircraft’s fuselage radius. The line representing the camera’s viewpoint is formulated as:
\begin{equation}
\vec{l} = \vec{x} + t\vec{v}
\label{eq:5}
\end{equation}
Where \begin{math}\vec{l}\end{math} represents the camera’s viewpoint, \begin{math}\vec{x}\end{math} is the camera's position, \begin{math}\vec{v}\end{math} is obtained by rotating the camera’s default direction vector by quaternion \begin{math}q\end{math}, and \begin{math}t\end{math} is a variable that determines the position of any point along line \begin{math}\vec{l}\end{math}.

For every pair of camera position and orientation, we use equations \eqref{eq:4} and \eqref{eq:5} to solve for \begin{math}t\end{math} where \begin{math}c = \vec{l}\end{math} to determine the point of intersection between line \begin{math}\vec{l}\end{math} and the surface of the cylinder. Since a line may intersect the surface of a cylinder at up to two points, only the point nearest to the camera’s position, \begin{math}x\end{math}, is kept. Fig.~\ref{fig:6} illustrates how the aircraft fuselage’s surface is modelled as the surface of a cylinder, as well as how \begin{math}\mathcal{L}_x\end{math} and \begin{math}\mathcal{L}_q\end{math} can be related by \begin{math}\mathcal{L}_c\end{math}. We combine our proposed loss component with \eqref{eq:2} to result in:

\begin{figure}[t]
\centerline{\includegraphics[width=\columnwidth]{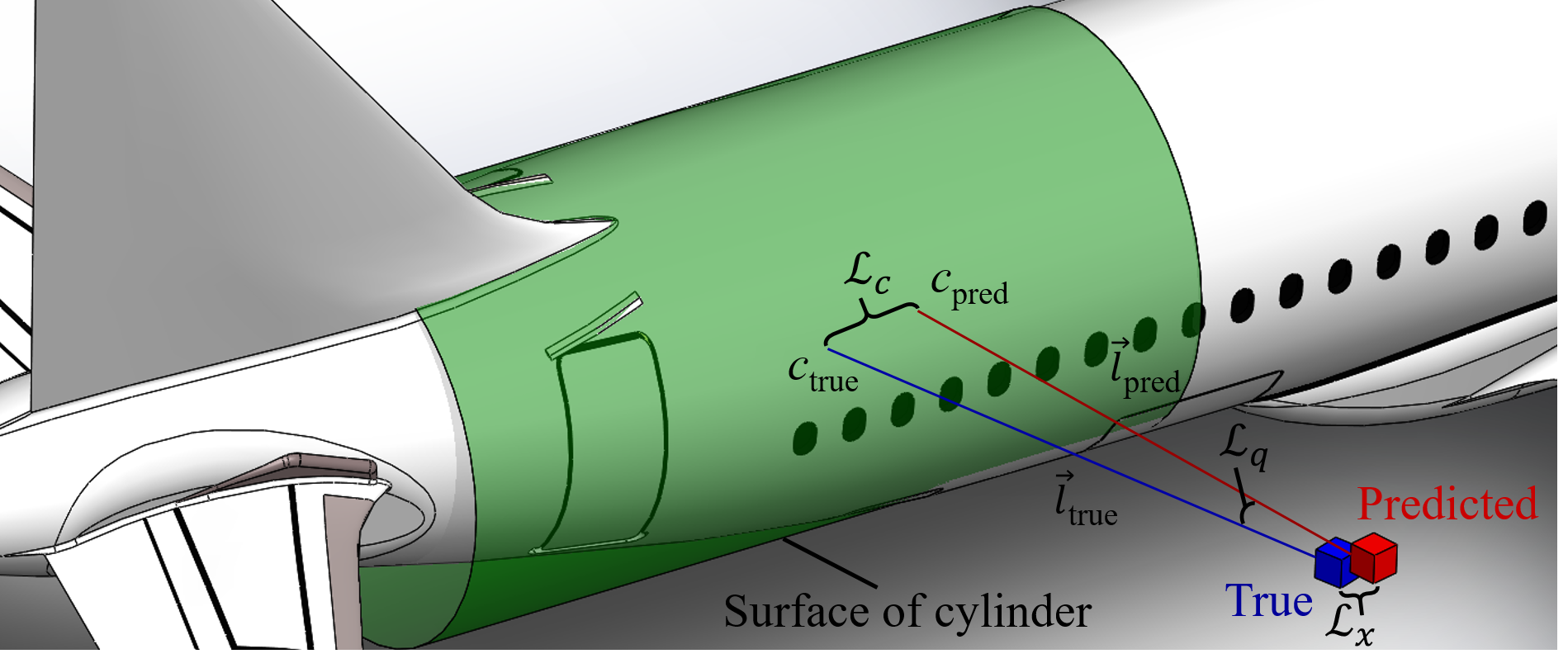}}
\vspace{-1mm}
\caption{Visualisation of how the top half of an aircraft’s fuselage is modelled as the surface of a cylinder (green). Losses \begin{math}\mathcal{L}_x\end{math}, \begin{math}\mathcal{L}_q\end{math} and \begin{math}\mathcal{L}_c\end{math} can also be visualised as the difference between their respective true and predicted components.}
\label{fig:6}
\vspace{-6mm}
\end{figure}

\begin{multline}
\mathcal{L}_\sigma(I) =  \mathcal{L}_x(I)\hat{\sigma}^{-2}_x +\log\hat{\sigma}^2_x + \mathcal{L}_q(I)\hat{\sigma}^{-2}_q
+\log\hat{\sigma}^2_q \\ + \mathcal{L}_c(I)\hat{\sigma}^{-2}_c +\log\hat{\sigma}^2_c
\label{eq:6}
\end{multline}
Where \begin{math}\mathcal{L}_c(I) = \|c-\hat{c}\|_2\end{math}, and \begin{math}c-\hat{c}\end{math} represent the difference between the true and predicted point of intersection coordinates. While the variance \begin{math}\sigma^{2}\end{math} is learnt, the logarithmic regularisation term prevents the network from learning an infinite variance to achieve zero loss. Hence, \begin{math}\hat{s}:=\log\hat{\sigma}^2\end{math} is learnt during implementation as it avoids a potential division by zero, resulting in the following function:

\begin{multline}
\mathcal{L}_\sigma(I) =  \mathcal{L}_x(I)\exp(-\hat{s}_x) + \hat{s}_x + \mathcal{L}_q(I)\exp(-\hat{s}_q) + \hat{s}_q \\
+ \mathcal{L}_c(I)\exp(-\hat{s}_c) + \hat{s}_c
\label{eq:7}
\end{multline}
Where \begin{math}\hat{s}_x\end{math}, \begin{math}\hat{s}_q\end{math} and \begin{math}\hat{s}_c\end{math} are learnt and we arbitrarily initialise all of them to zero. We refer to our proposed additional loss component as the Image Centre Scene Coordinate (ICSC) loss and the network with loss function \eqref{eq:7} as ICSC-PoseNet.

\section{Scan Path Generation and Image Localisation} \label{scan_path}

After estimating the PTZ camera's pose, we propose three steps to generate a path in pan-tilt values for the camera to follow to scan the upper half of the aircraft. The first step extracts relevant scan sections from within the point cloud of the 3D aircraft model and performing a linear interpolation on each sections separately. Secondly, the interpolated surface points are converted from Cartesian coordinates into an array of pan-tilt values relative to the estimated PTZ camera's pose. Lastly, an algorithm extracts scan points from the array of pan-tilt values based on the desired FOV and amount of overlap between captured scan images.

\subsection{Interpolation of Aircraft Model Point Cloud}
\begin{figure}[t]
\centerline{\includegraphics[width=\columnwidth]{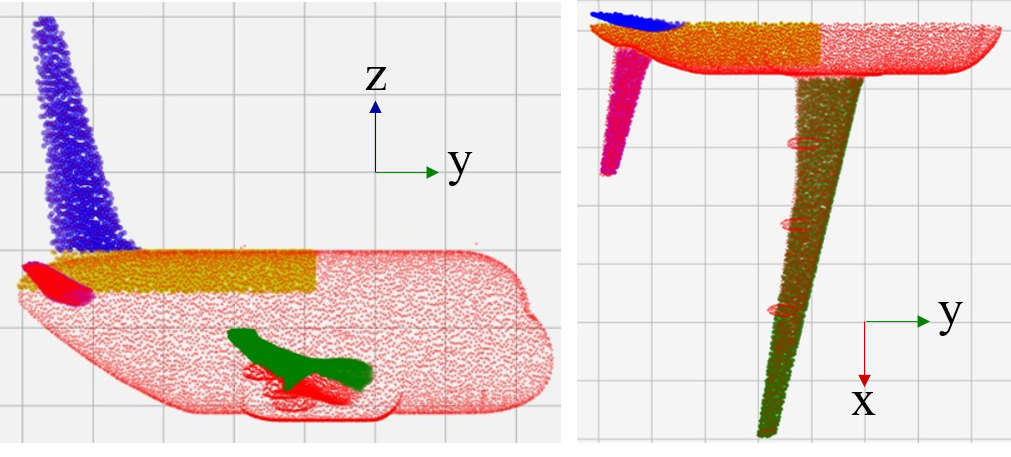}}
\vspace{-2mm}
\caption{Side view (left) and top view (right) of point cloud with direction of axes. Example of sectioning point cloud to fuselage (yellow), tail (blue), wing (green) and stabiliser (purple) for scanning by a camera in quadrant 3 are shown.}
\label{fig:70}
\vspace{-3mm}
\end{figure}

Interpolation of the point cloud (from the 3D aircraft model) is performed separately on different features (i.e. the fuselage, tail, wing and horizontal stabiliser) as their major surfaces lie on different planes. We define the relevant scan sections for a camera positioned in the back half (quadrants 2 and 3) of the aircraft as the tail, top half of the wing, the stabiliser and the back half of the fuselage, while a camera in the front half (quadrants 1 and 4) will scan the remaining front half of the fuselage. Only the upper-half of the fuselage (above the windows) are included in the scan. Fig.~\ref{fig:70} shows the side and top view of the aircraft model's point cloud with different scan sections relevant to quadrant 3 coloured for visualisation. For points on the fuselage, stabiliser and wing, z-coordinates are interpolated over x and y while for points on the tail, their x-coordinates are interpolated over y and z. All interpolation are performed at 5 cm intervals using SciPy's griddata function \cite{2020SciPy-NMeth} to form an array for each section where each row i has the same x-coordinate (except for the tail where each row has the same z-coordinate) and each column j has the same y-coordinate as such:

\begin{equation}
\begin{bmatrix}
(x_{11}, y_{11}, z_{11}) & (x_{12}, y_{12}, z_{12}) & \hdots & (x_{1j}, y_{1j}, z_{1j})\\
(x_{21}, y_{21}, z_{21}) & (x_{22}, y_{22}, z_{22}) & \hdots & (x_{2j}, y_{2j}, z_{2j})\\
\vdots & \vdots & \ddots & \vdots\\
(x_{i1}, y_{i1}, z_{i1}) & (x_{i2}, y_{i2}, z_{i2}) & \hdots & (x_{ij}, y_{ij}, z_{ij})\\
\end{bmatrix}
\label{array:xyz}
\end{equation}

\subsection{Point Cloud to Camera Pan-tilt}
To convert the points in the interpolated point cloud into camera pan-tilt values, the camera is first placed at its estimated position with respect to the aircraft in the Cartesian space. Next, every point in the point cloud is converted into their respective pan-tilt values with the assumption that both the camera's and aircraft's x-y plane are parallel to each other. This is achieved by relating the camera's pan and tilt angles to the same angles (\begin{math}\varphi\end{math} and \begin{math}\theta\end{math}) of a spherical coordinate system and can be converted from the Cartesian coordinate system in a similar manner. The equations for converting every point into camera pan and tilt are:

\begin{equation}
\varphi = \arctan\frac{y_c}{x_c} - \alpha
\label{eq:pan}
\end{equation}

\begin{equation}
\theta = \arctan\frac{\sqrt{(x_c^2 + y_c^2)}}{z_c}
\label{eq:tilt}
\end{equation}

Where \begin{math}\varphi\end{math} and \begin{math}\theta\end{math} are pan and tilt respectively, while \begin{math}x_c\end{math}, \begin{math}y_c\end{math}, and \begin{math}z_c\end{math} are the displacements of each point in the interpolated point cloud with respect to the PTZ camera in the x, y, and z axis respectively. \begin{math}\alpha\end{math} represents the yaw error which is expected to be within ${\pm} 10${\degree} as described in section \ref{proposed_setup}, and can be obtained using:

\begin{equation}
\alpha = \gamma - \beta_q
\label{eq:alpha}
\end{equation}

Where \begin{math}\gamma\end{math} is the camera's estimated angle in the yaw direction (rotation about the z-axis) extracted from the CNN's predicted quarternion output, with 0{\degree} being the yaw angle if the camera is facing exactly perpendicular to the aircraft's fuselage. \begin{math}\beta_q\end{math} is the pre-defined angle rotated by the camera for initialisation depending on the quadrant, q, it is placed in: \begin{math}\beta_1 = +10{\degree}; \beta_2 = -20{\degree}; \beta_3 = +20{\degree}; \beta_4 = -10{\degree}\end{math}.

Equations \eqref{eq:pan}, \eqref{eq:tilt} and \eqref{eq:alpha} are used to compute the respective pan-tilt values for every valid element in the array, \eqref{array:xyz} of each section, forming another array, $U$, with a similar organisation:

\begin{equation}
U = 
\begin{bmatrix}
(p_{11}, t_{11}) & (p_{12}, t_{12}) & \hdots & (p_{1j}, t_{1j})\\
(p_{21}, t_{21}) & (p_{22}, t_{22}) & \hdots & (p_{2j}, t_{2j})\\
\vdots & \vdots & \ddots & \vdots\\
(p_{i1}, t_{i1}) & (p_{i2}, t_{i2}) & \hdots & (p_{ij}, t_{ij})\\
\end{bmatrix}
\label{array:pt}
\end{equation}

Where \begin{math}p_{ij}\end{math} and \begin{math}t_{ij}\end{math} are the computed camera pan and tilt values, respective to each element with coordinates \begin{math}(x_{ij}, y_{ij}, z_{ij})\end{math} in \eqref{array:xyz}.

\begin{algorithm}[b]
\caption{Generate PTZ Camera Scan Path}
\begin{algorithmic}[1]
\STATE\textbf{Input: }{Array $U$}
\STATE\textbf{Output: }{List $V$ containing appended scan points}
\STATE Define $\lambda = 1 - \mu$ (overlap factor)
\STATE $m_{last}$ = $median(t_{1,all})+\lambda\times VFOV$
\FOR{row $i$ in $U$}
    \STATE $m_{next}$ = $median(t_{i,all})$
    \STATE $n_{last}$ = $p_{i1}+\lambda\times HFOV$
    \IF{$|m_{last}-m_{next}|\geq(\lambda\times VFOV)$ \OR $i$ is last row of $U$ \AND $|m_{last} - m_{next}| > VFOV/2$}
        \STATE $m_{last} = m_{next}$
        \FOR{col $j$ in row $i$}
            \STATE $n_{next}$ = $p_{ij}$
            \IF{$|n_{last}-n_{next}|\geq(\lambda\times HFOV)$ \OR $j$ is last col of $i$ \AND $|n_{last} - n_{next}| > HFOV/2$}
                \STATE Append ($p_{ij}$, $t_{ij}$) to $V$
                \STATE $n_{last} = n_{next}$
            \ENDIF
        \ENDFOR
    \ENDIF
\ENDFOR
\end{algorithmic}
\label{alg:algo1}
\end{algorithm}

\subsection{Scan Path Generation and Image Localisation}
To reduce the time taken to scan the aircraft, we generate an efficient scan path that does not capture images with unnecessary amount of overlap. This is achieved by taking into account both the desired Hoziontal FOV (HFOV) and Vertical FOV (VFOV) of the camera during the scan, as well as an overlap factor, \begin{math}\mu\end{math}, which represents the desired minimum ratio of overlap between two consecutive images in terms of their HFOV and VFOV. We choose HFOV and VFOV to be 6.15{\degree} and 3.46{\degree} respectively (13x zoom) for our scans, and \begin{math}\mu\end{math} to be 0.15 as we observe that it results in sufficient overlaps between images.

We propose algorithm \ref{alg:algo1} (for fuselage and tail sections) to generate a list of scan points, $V$, given array $U$ \eqref{array:pt} for each section and a pre-defined overlap factor \begin{math}\mu\end{math}. For the fuselage and tail, the separation between rows in array $U$ is related to the variation in camera tilt and VFOV. On the other hand, the separation between rows for array $U$ of the wing and stabiliser is related to the camera pan and HFOV. Hence, the algorithm for the wing and stabiliser is similar except that all \begin{math}p_{ij}\end{math} are replaced by \begin{math}t_{ij}\end{math} and vice versa, and all VFOV are replaced by HFOV and vice versa. The eventual scan path for each section is in the order of the pan-tilt values in list $V$. Each scan image, captured from \begin{math}(p_{ij}, t_{ij})\end{math}, is labelled with its Cartesian coordinate by retrieving its respective coordinate \begin{math}(x_{ij}, y_{ij}, z_{ij})\end{math} from \eqref{array:xyz}. This coordinate represents the estimated location of the image's centre pixel on the surface of the aircraft.

\section{Experimental Study}
\subsection{Experimental Setup}
\begin{figure}[t]
\centerline{\includegraphics[width=0.9\columnwidth]{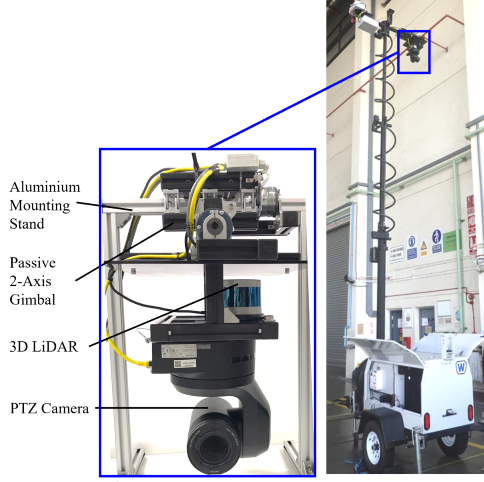}}
\vspace{-2mm}
\caption{(Left) prototype: a 3D LiDAR and a PTZ camera mounted onto a passive 2-axis gimbal and secured onto a mounting stand during lab tests. (Right) the same prototype gimbal mounted to the top of a mast extended to about 7 m from the ground.}
\label{fig:setup}
\vspace{-6mm}
\end{figure}

To evaluate our approach, we requested and obtained special access to an A320 to gather images and pose-related data for our real test dataset. We built a prototype consisting of a Panasonic AW-UE150 PTZ Camera and a Velodyne VLP-16 3D LiDAR mounted onto a passive 2-axis gimbal. The prototype was secured onto the top of a vertical mast that is extendable up to 8 m from the ground and brought within the proposed boundary. Fig.~\ref{fig:setup} shows the prototype in the lab and when mounted onto the mast. Images for the entire setup next to the aircraft are not shown as they are deemed sensitive by the venue and airline. The 3D LiDAR was only used to obtain ground truth, requiring multiple minor manual adjustments to ensure the desired features are observed within the point cloud, and is not used in the proposed methodology.

\subsection{Real Dataset and Aircraft Scan Images}
Real images with their ground truth poses were gathered from three scenes, each containing an A320 with different paintwork, background and lighting to evaluate our method. We include results from our previous work \cite{CPE_Aircraft} (Real Scene 1 - captured at 60{\degree} HFOV using a different PTZ camera) as they are relevant and gathered new real data captured at 72.5{\degree} HFOV. All images were obtained from 16 different positions in quadrant 3 by first following steps 1 and 2 in section \ref{proposed_setup} to deploy the camera (with some intentionally nearer to the border to simulate less-careful positioning). To obtain variations of camera base orientation, steps 3 and 4 of section \ref{proposed_setup} were then repeated four to six times while keeping the camera base's position unchanged and an image was captured after each repetition. A total of 73 real images were captured from unique camera poses. To illustrate the ability of our scan path planning method described in section \ref{scan_path}, we used one of the camera poses predicted by ICSC-PoseNet using a real image captured from quadrant 3 as input to generate its scan path and capture scan images after initialisation.

\newcommand\boldblue[1]{\textcolor{blue}{\textbf{#1}}}

\begin{table*}[t]
\centering
 \begin{tabular}{c c c c c c} 
 \hline
 \centering Expt. (HFOV) & Fine-tune on (images) & Test on (images) & Error & PoseNet+\cite{R18} & ICSC-PoseNet (Ours) \\ 
 \hline
 \multirow{2}{5em}{\centering 1 (60{\degree})} & \multirow{2}{10em}{\centering Syn60 (4000)} & 
 \multirow{2}{10em}{\centering Syn60 (300)} & 
 \centering Median & 0.067m, 0.60{\degree} & \boldblue{0.063m, 0.50{\degree}} \\
 & & & \centering RMSE & 0.10m, 0.75{\degree} & \boldblue{0.085m, 0.63{\degree}} \\
 \multirow{2}{5em}{\centering 2 (72.5{\degree})} & \multirow{2}{10em}{\centering Syn725 (4000)} & 
 \multirow{2}{10em}{\centering Syn725 (300)} & 
 \centering Median & 0.12m, 0.56{\degree} & \boldblue{0.050m, 0.36{\degree}} \\
 & & & \centering RMSE & 0.12m, 0.63{\degree} & \boldblue{0.065m, 0.39{\degree}} \\

 \hline
 \multirow{2}{5em}{\centering 3 (60{\degree})} & \multirow{2}{10em}{\centering Syn60 (4000)} & 
 \multirow{2}{10em}{\centering Real Scene 1 (28)} & 
 Median & 0.292m, 1.25{\degree} & \boldblue{0.217m, 0.73{\degree}} \\ 
 & & & RMSE & 0.312m, 1.44{\degree} & \boldblue{0.237m, 0.88{\degree}} \\
 \multirow{2}{5em}{\centering 4 (72.5{\degree})} & \multirow{2}{10em}{\centering Syn725 (4000)} & 
 \multirow{2}{10em}{\centering Real Scene 2 (15)} & 
 Median & 0.20m, 2.51{\degree} & \boldblue{0.14m, 1.78{\degree}} \\ 
 & & & RMSE & 0.267m, 2.742{\degree} & \boldblue{0.16m, 1.94{\degree}} \\
 \multirow{2}{5em}{\centering 5 (72.5{\degree})} & \multirow{2}{10em}{\centering Syn725 (4000)} & 
 \multirow{2}{10em}{\centering Real Scene 3 (30)} & 
 Median & \boldblue{0.16m}, 1.26{\degree} & 0.18m, \boldblue{1.09{\degree}} \\ 
 & & & RMSE & \boldblue{0.19m}, 1.40{\degree} & 0.20m, \boldblue{1.13{\degree}} \\

 \hline
 \end{tabular}
 \caption{Comparison of our proposed ICSC-PoseNet against PoseNet+ \cite{R18} for all scenes when fine-tuned and tested on different synthetic (Syn60, Syn725) or real (Real Scene 1,2,3) images, and with different camera HFOV (60{\degree}, 72.5{\degree}). Each Real Scene contains an A320 with different paintwork, background and lighting. Their performance are evaluated by median and root-mean-square error (RMSE) of position and orientation with the lowest error for each scene in bold and blue. Our proposed method achieves lower error when tested on all synthetic and real scenes, except for Real Scene 3 where the difference in position error is marginal. For reference, the distance between numbered frames of an A320, used by operators to locate defects on the fuselage, is about 0.53 m and improving localisation error helps reduce the chance that detected defects are wrongly tagged to an adjacent frame.}
 \vspace{-5mm}
\label{tab1}
\end{table*}

\subsection{Network Implementation and Experiments}
We implemented two networks (PoseNet+ \cite{R18} as baseline and our ICSC-PoseNet) using TensorFlow, supported by a NVIDA RTX Turbo 2080Ti GPU. We fine-tuned the networks pre-trained on ImageNet \cite{R36} to leverage on transfer learning and used Xception \cite{R35} (a state-of-the-art architecture) as their base architecture. All input images are normalised such that all pixel intensities range from -1 to 1. We optimised both networks with ADAM \cite{R37} using default parameters at a learn rate of \begin{math}10^{-4}\end{math} and a batch size of 25. We adapted the implementation \cite{multitasklosscode} of multi-task loss introduced by Kendall \textit{et al.} \cite{multitaskloss} for PoseNet+'s and our proposed loss function. To compare performance of the two networks, each network was separately fine-tuned using synthetic images generated at 60{\degree} (Syn60) or 72.5{\degree} (Syn725) HFOV and tested on synthetic and real images obtained with the same camera HFOV. To evaluate fine-tuning on a small set of real images, we also fine-tuned each network on real images extracted from the real dataset and tested on the remaining real images not used during fine-tuning. Each network was fine-tuned (from the pre-trained weights) several times between 50-200 epochs on the synthetic datasets, and 200 epochs on the real dataset. We use the best performance of each network for evaluation and present all results in Table~\ref{tab1} with their experimental details.

\section{Results and Discussion}

We discuss the performance of our proposed method in five aspects: (a) deploying a PTZ camera within proposed boundaries; (b) sim-to-real camera pose estimation; (c) improvements due to geometric loss component; (d) large synthetic dataset vs small real dataset; and (e) scan path planning and localisation of scan images to demonstrate our proposed workflow. Table~\ref{tab1} shows the results from our proposed network when fine-tuned and tested on both synthetic and real datasets, with a comparison with PoseNet+ \cite{R18}. 

\subsection{Deploying a PTZ Camera Within Proposed Boundaries}
\begin{figure}[t]
\centering
\centerline{\includegraphics[width=0.9\columnwidth]{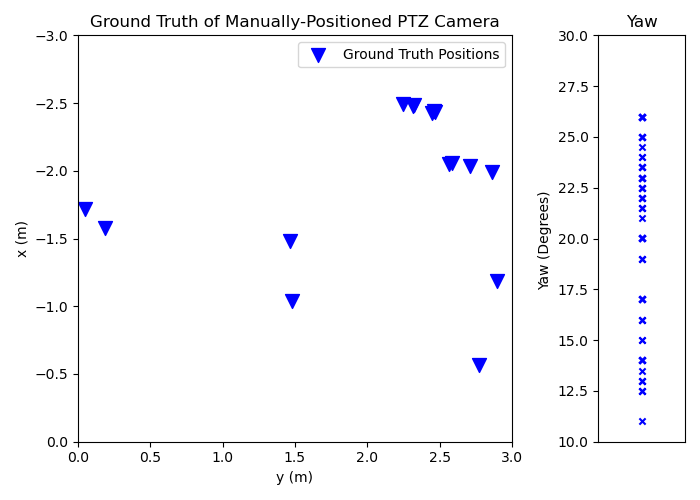}}
\vspace{-3mm}
\caption{Spread of ground truth 2D positions (left) and orientations (right) of real images captured by the PTZ camera at a height (z) of between 6.25-7.25 m, following steps proposed in section \ref{proposed_setup}. All ground truth lie within our proposed boundaries (3 m x 3 m and 10{\degree}-30{\degree}), demonstrating that the proposed use of aircraft features as landmarks to guide the initial manual set up of the PTZ camera within these boundaries is feasible. A total of 73 images with unique combinations of position and orientation are collected and used for evaluating localisation accuracy.}
\label{fig:gtpositionyaw}
\vspace{-5mm}
\end{figure}

All real images are obtained from manually setting up the PTZ camera using the steps proposed in section \ref{proposed_setup}. Fig.~\ref{fig:gtpositionyaw} shows the spread of ground truth of all real images obtained and we find that they all lie within our proposed boundary, suggesting that the proposed use of aircraft features (windows and pylons) as landmarks to guide the initial manual set up of the PTZ camera's position and orientation is feasible. This is an important step since the use of DCNN for CPE performs best within a pre-defined range of predictions (for both position and orientation) included in the fine-tuning dataset, as deep pose estimators underperform in the task of generalising to unseen scenes \cite{R9}. However, this is only useful under the assumption that consistent deployment of the PTZ camera within the proposed boundary is feasible in the first place. We show that we can manually position and orientate our PTZ camera within the same proposed boundary and pan range used to generate our synthetic images.

\begin{figure}[t]
\centering
\centerline{\includegraphics[width=\columnwidth]{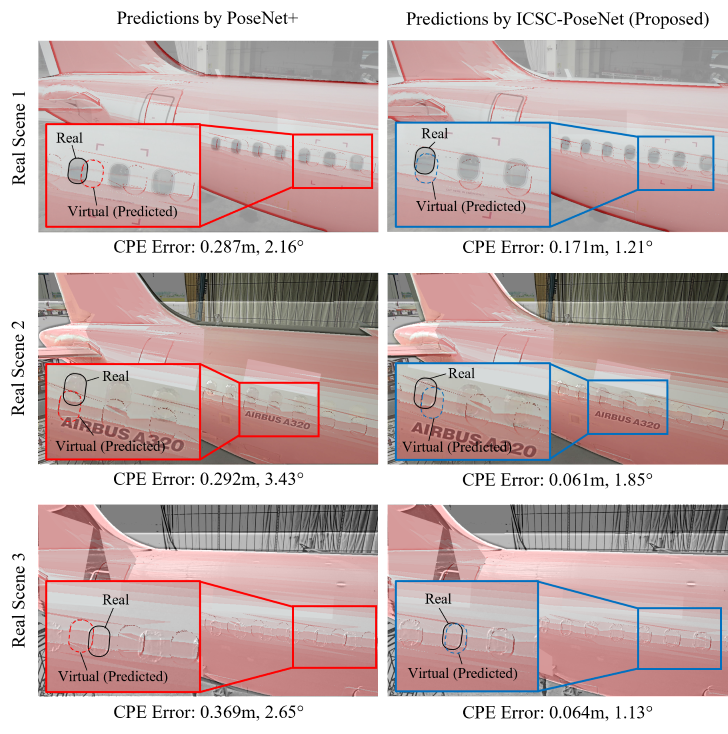}}
\vspace{-3mm}
\caption{Example predictions extracted from different scenes, each with an A320 with different paintwork, background and lighting. Features that are sensitive in nature are covered to protect the airlines' privacy. Virtual images captured from predicted camera poses are overlaid in red onto their respective real input image. ICSC-PoseNet (proposed) achieves lower camera pose estimation (CPE) error and results in better overlap compared to those by PoseNet+ \cite{R18} (larger offsets, more visible in the enlarged regions).}
\label{fig:overlayss}
\vspace{-6mm}
\end{figure}

\begin{figure*}[t]
\centering
\centerline{\includegraphics[width=\textwidth]{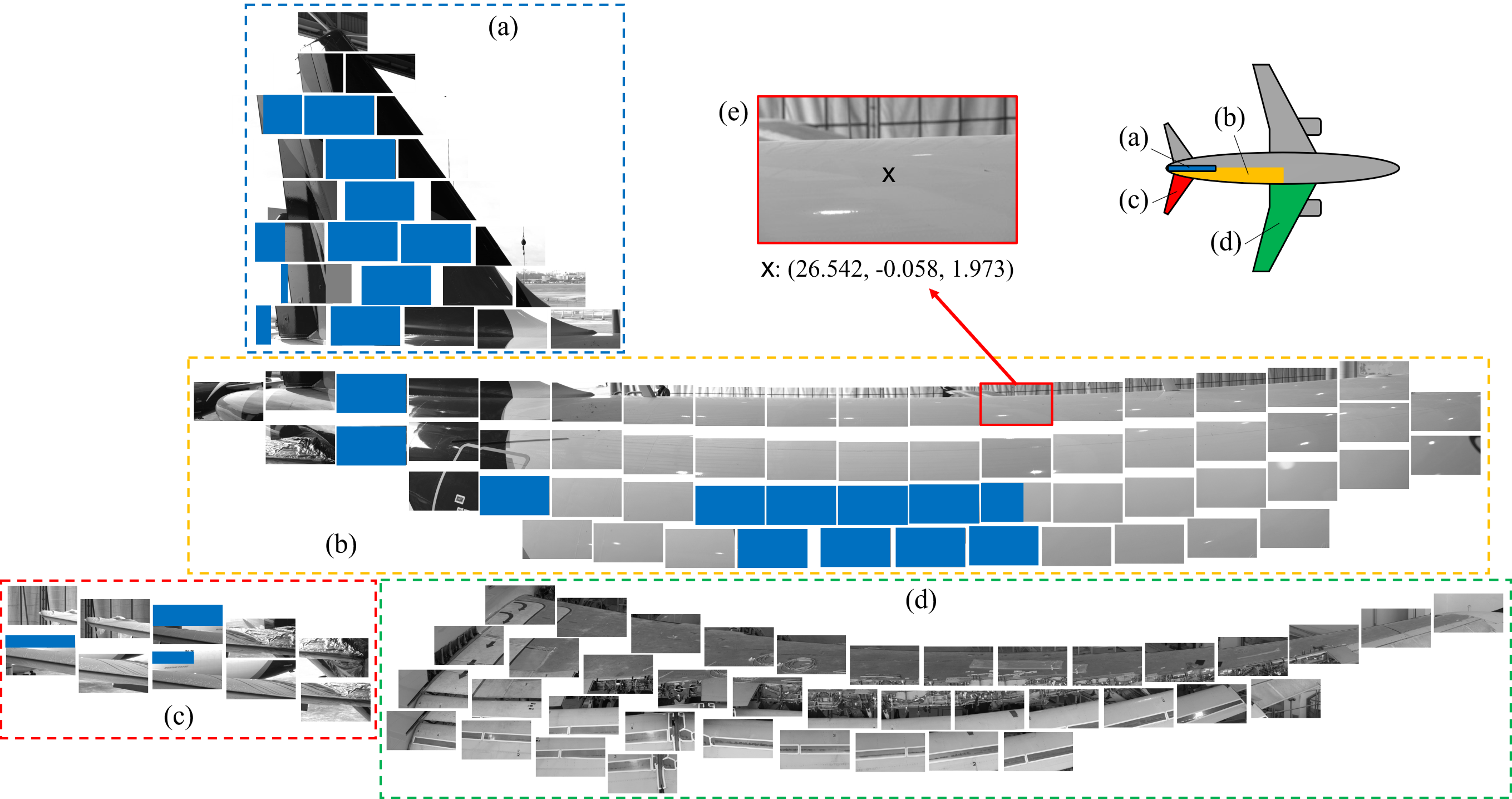}}
\vspace{-1mm}
\caption{A full set of scan images captured from quadrant 3 at 10X zoom using our proposed workflow, after initialisation using camera pose estimation via ICSC-PoseNet. All images have been converted to greyscale and certain parts of the image have been covered in blue to protect the airline's privacy. The images are organised into a montage and grouped into four sections: (a) tail, (b) fuselage, (c) stabiliser, and (d) wing (in a partially-opened state for maintenance). All scan images are localised as part of our workflow and labelled with their centre-pixel's Cartesian coordinates w.r.t. an aircraft coordinate system. (e) is an example of a scan image with its position label.}
\label{fig:montage_q3}
\vspace{-6mm}
\end{figure*}

\subsection{Sim-to-real Camera Pose Estimation}
Our network is able to estimate a PTZ camera’s pose relative to an Airbus A320 using a single image as input and without fine-tuning on any real images. This is achieved without any knowledge of the scene other than the aircraft's 3D model, and without any additional infrastructure. Our network (ICSC-PoseNet), when fine-tuned on only synthetic images, obtains median and RMSE prediction errors of less than 0.24 m and 2{\degree} across all real scenes (Table~\ref{tab1}) which is sufficient for initialisation given the scale of the aircraft. Interestingly, fine-tuning PoseNet+ using a limited set of 36 real images from Real Scene 1 \& 2 results in higher median and RMSE error, both 0.50m and about 7{\degree}, when tested on 9 unseen real images from the same scenes. This further supports the use of synthetic images over real images in this application, especially given the difficulty of obtaining real images in the first place. Fig.~\ref{fig:overlayss} shows sample synthetic images captured from their predicted camera poses overlaid onto their respective real input images and we observe generally good overlap of aircraft outline. Our results show the network’s ability to extract relevant aircraft features from the randomised textures in the synthetic dataset and match their scale and position with the real input images to regress a camera pose, demonstrating successful sim-to-real transfer for this task.

\subsection{Improvements Due to Geometric Loss Component}
With reference to Table~\ref{tab1}, our network achieves lower median errors and RMSE as compared to PoseNet+ for both Experiments 1 \& 2 that test on only synthetic images at different HFOV. As the synthetic images used for fine-tuning and testing belong to the same domain, the results show that our proposed loss function improves camera pose estimation when there is no domain gap. Our network also out-performs PoseNet+ when tested on all real scenes, except in Real Scene 3 where the difference in their position errors is marginal (about 10 percent) and argue that it is likely due to sim-to-real differences of that scene. Example predictions in Fig.~\ref{fig:overlayss} demonstrate that ICSC-PoseNet improves CPE accuracy and achieves smaller offsets between real and predicted positions of aircraft features. For comparison, an A320's window-to-window distance is about 0.53 m, which is also the distance between frame numbers used by operators to locate features and defects on the fuselage. Any reduction in localisation error is beneficial for defect detection systems as it reduces the chance that a detected defect is wrongly tagged to an adjacent frame. We conclude that our proposed additional component in the loss function which geometrically relates the predicted position and orientation improves camera pose estimation accuracy in our task.

\subsection{Scan Path Planning and Localisation of Scan Images}
To demonstrate the result of our scan and the successful completion of our proposed workflow, a total of 134 scan images are captured at 10x zoom from one instance of CNN-based initialisation within quadrant 3. We group the scan images into their respective sections (i.e. tail, fuselage, stabiliser, and wing) and organise them into a montage shown in Fig.~\ref{fig:montage_q3}. As intended by our algorithm, we observe a small overlap between images to reduce the chance of missing areas while keeping the total scan images low for efficiency. Every image was captured from a computed set of camera pan and tilt values and labelled with their respective Cartesian coordinates on the aircraft's surface (retrieved from their respective elements \begin{math}(x_{ij}, y_{ij}, z_{ij})\end{math} in \eqref{array:xyz}). This achieves localisation of every scan image and can be used to localise and label surface defects during visual inspection tasks.

\section{Conclusion}
We demonstrate camera pose estimation with respect to an aircraft, as well as a workflow for scan path generation and image localisation. Unlike existing methods, this is easy to deploy and achieved without additional infrastructure, physical contact with or prior access to a real aircraft. We are successful in sim-to-real transfer for our CPE task, by fine-tuning a DCNN with synthetic images generated with domain randomisation. We also show that providing a geometric relationship between the predicted position and orientation as an additional component in the loss function can improve pose estimation performance and the localisation of aircraft features in images. Future work can explore improving the initial pose estimate by using multiple input images captured from the same PTZ camera to benefit from spatio-temporal information, as well as sensor fusion with other sensor data such as from a LiDAR.

\section*{Acknowledgment}
This research is supported by ST Engineering Aerospace as part of a project with the Civil Aviation Authority of Singapore to develop a GVI system for detecting damage to the exterior of aircraft due to lightning strikes.

\bibliographystyle{./bibliography/IEEEtran}
\bibliography{./bibliography/main}

\begin{IEEEbiography}[{\includegraphics[width=1in,height=1.25in,clip,keepaspectratio]{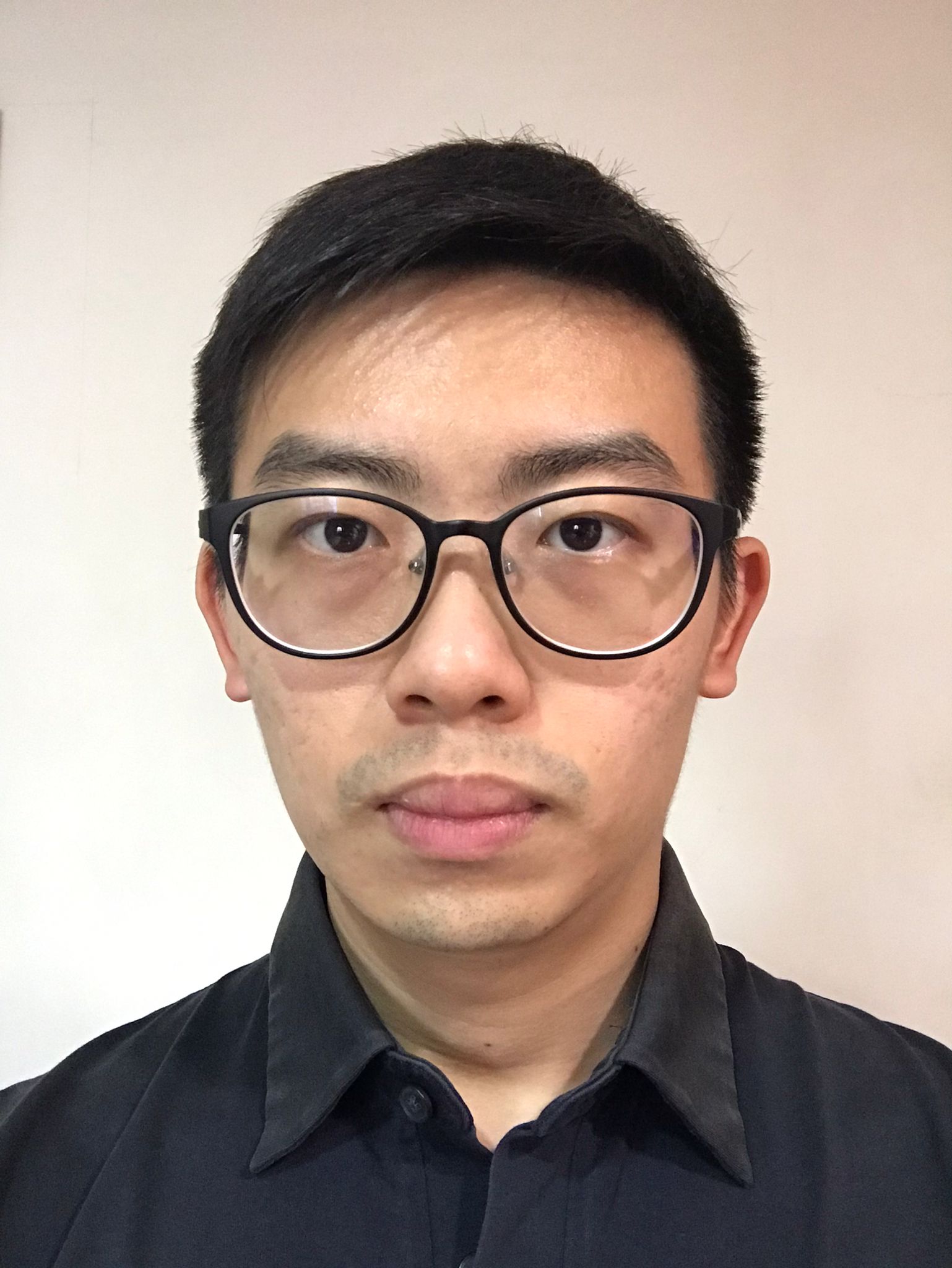}}]{Xueyan Oh}
received the B.Eng. (Hons.) degree in engineering from the Singapore University of Technology and Design (SUTD), Singapore, in 2016. He is currently working toward the Ph.D. degree in Engineering Product Development (EPD) with the SUTD Engineering Product Development (EPD) Pillar, Singapore. His research interests include using deep learning for easily deployable vision-based localisation methods.
\end{IEEEbiography}

\begin{IEEEbiography}[{\includegraphics[width=1in,height=1.25in,clip,keepaspectratio]{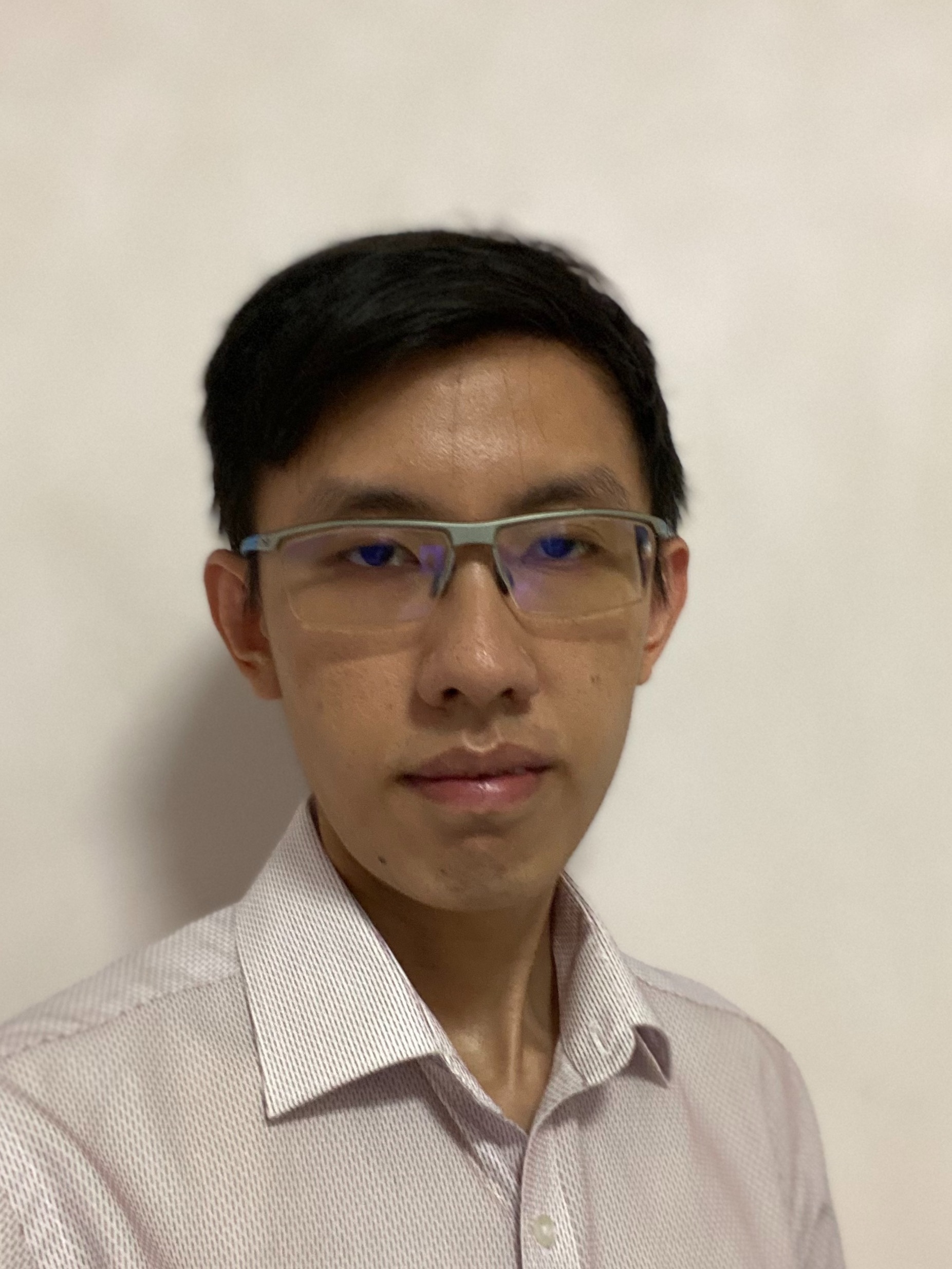}}]{Leonard Loh}
received the M.Tech. in Intelligent Systems from the National University of Singapore, Singapore, in 2022. He is a Research Assistant with the Singapore University of Technology and Design. His work involves software development \& integration for various projects, such as an automated defect detection system. He is also working on developing a map merging algorithm for multiple robot exploration. 
\end{IEEEbiography}

\begin{IEEEbiography}[{\includegraphics[width=1in,height=1.25in,clip,keepaspectratio]{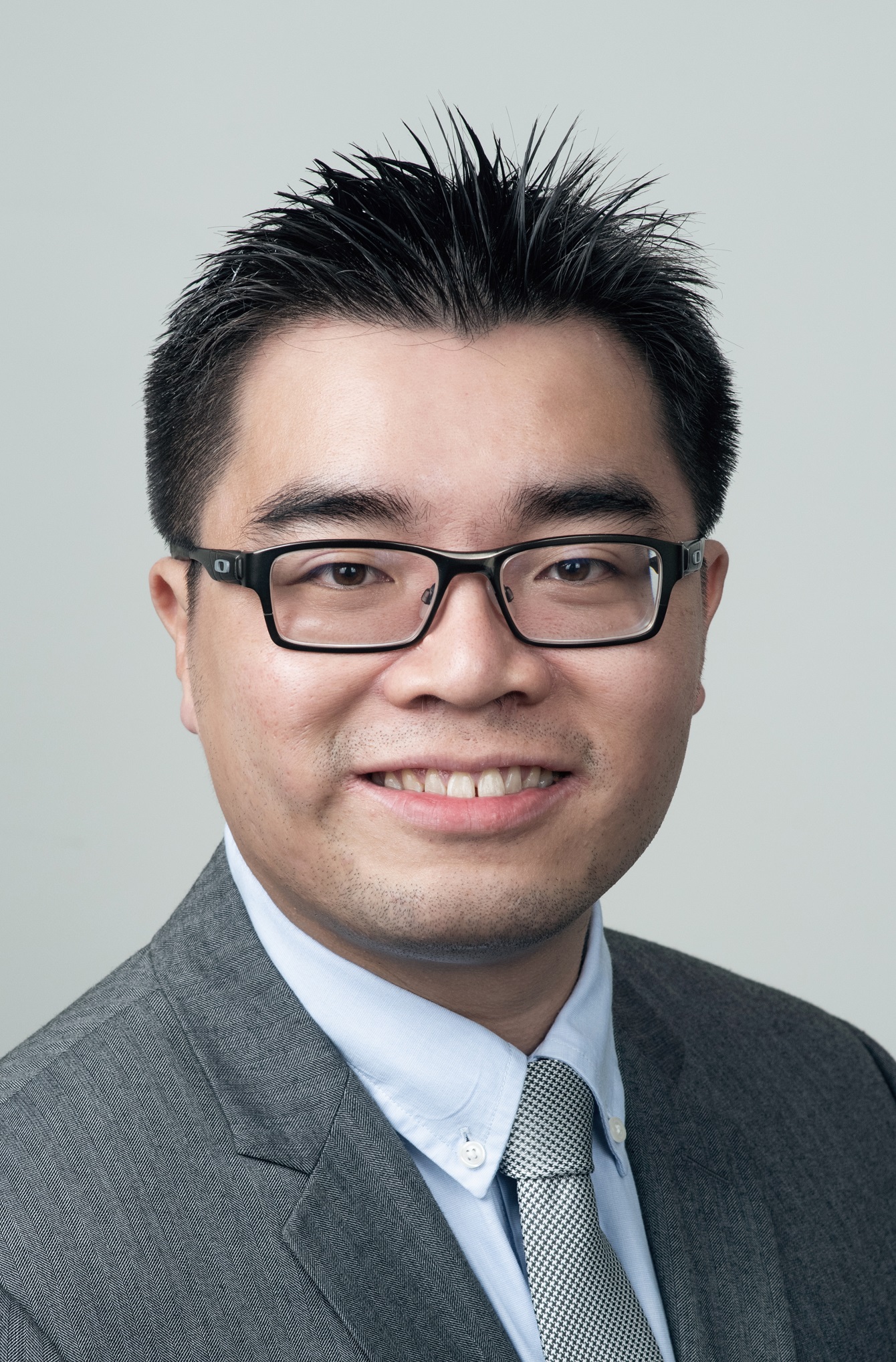}}]{Shaohui Foong}
is an Associate Professor in the Engineering Product Development (EPD) Pillar at the Singapore University of Technology and Design (SUTD) and Senior Visiting Academician at the Changi General Hospital, Singapore. He received his B.S., M.S. and Ph.D. degrees in Mechanical Engineering from the George W. Woodruff School of Mechanical Engineering, Georgia Institute of Technology, Atlanta, USA. In 2011, he was a Visiting Assistant Professor at the Massachusetts Institute of Technology, Cambridge, USA. His research interests include system dynamics \& control, nature-inspired robotics, magnetic localization, medical devices and design education \& pedagogy.
\end{IEEEbiography}

\begin{IEEEbiography}[{\includegraphics[width=1in,height=1.25in,clip,keepaspectratio]{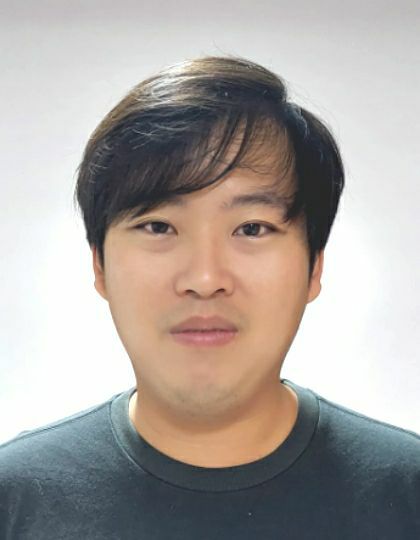}}]{Zhong Bao Andy Koh}
received the B.Eng. degree in Aerospace Engineering from the Nanyang Technological University, Singapore in 2014. He is currently working on UAV systems in ST Engineering Aerospace Ltd.
\end{IEEEbiography}

\begin{IEEEbiography}[{\includegraphics[width=1in,height=1.25in,clip,keepaspectratio]{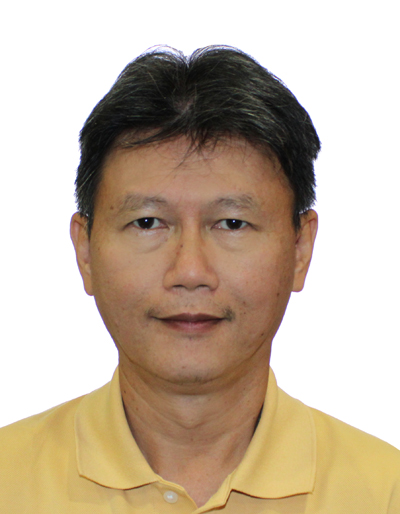}}]{Kow Leong Ng}
received the B.Sc. degree in Computing and Information Systems from the University of London, London, United Kingdom. He is currently a Software Engineer with ST Engineering Aerospace Ltd, Singapore.
\end{IEEEbiography}

\begin{IEEEbiography}[{\includegraphics[width=1in,height=1.25in,clip,keepaspectratio]{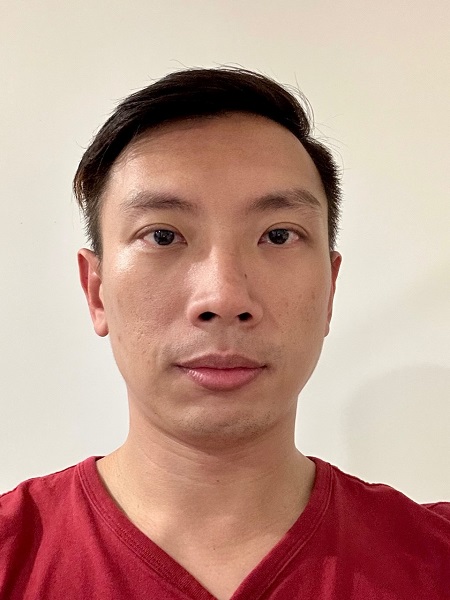}}]{Poh Kang Tan}
received the B.Eng. degree in Computer Engineering from Nanyang Technological University, Singapore. He is currently working as a Software Engineer in ST Engineering.
\end{IEEEbiography}

\begin{IEEEbiography}[{\includegraphics[width=1in,height=1.25in,clip,keepaspectratio]{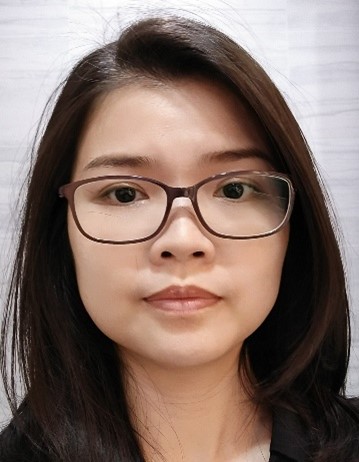}}]{Pei Lin Pearlin Toh}
received the B.Eng. degree in mechanical and production engineering from Nanyang Technological University, Singapore, in 2003. She is a PMP certified Programme Manager with ST Engineering Aerospace Ltd, specialising in Unmanned Air Systems solutions.
\end{IEEEbiography}

\begin{IEEEbiography}[{\includegraphics[width=1in,height=1.25in,clip,keepaspectratio]{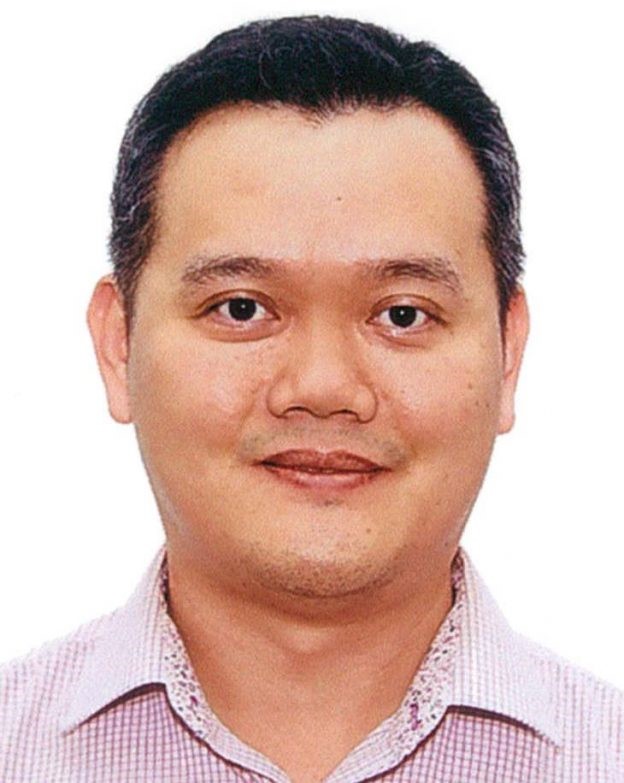}}]{U-Xuan Tan}
(Member, IEEE) received the B.Eng. and Ph.D. degrees from Nanyang Technological University, Singapore, in 2005 and 2010, respectively. From 2009 to 2011, he was a Postdoctoral Fellow with the University of Maryland, College Park, MD, USA. From 2012 to 2014, he was a Lecturer with the Singapore University of Technology and Design, Singapore, where he took up a research intensive role in 2014 and has been promoted to Associate Professor since 2021. He is also holding a Senior Visiting Academician position at Changi General Hospital. His research interests include mechatronics, on-site robotics algorithm, sensing and control, sensing and control technologies for human–robot interaction, and interdisciplinary teaching.
\end{IEEEbiography}


\end{document}